%% file: Unsupervised_deep_learning_by_injecting_low-rank_and_sparse_priors__v0.tex
\documentclass[conference]{IEEEtran}
\IEEEoverridecommandlockouts
\usepackage{cite}
\usepackage{amsmath,amssymb,amsfonts}
\usepackage{algorithmic}
\usepackage[dvipdfmx]{graphicx}
\usepackage{textcomp}
\usepackage{xcolor}
\usepackage{slashbox}
\def\BibTeX{{\rm B\kern-.05em{\sc i\kern-.025em b}\kern-.08em
    T\kern-.1667em\lower.7ex\hbox{E}\kern-.125emX}}
\begin{document}

\title{Unsupervised Deep Learning by\\ Injecting Low-Rank and Sparse Priors}

\author{\IEEEauthorblockN{1\textsuperscript{st} Tomoya Sakai}
\IEEEauthorblockA{\textit{School of Information and Data Sciences} \\
\textit{Nagasaki University}\\
Nagasaki, Japan}
}

\maketitle

\begin{abstract}
What if deep neural networks can learn from sparsity-inducing priors? 
When the networks are designed by combining layer modules (CNN, RNN, etc), 
engineers less exploit the inductive bias, i.e., existing well-known rules or prior knowledge,
other than annotated training data sets. 
We focus on employing sparsity-inducing priors in deep learning 
to encourage the network to concisely capture the nature of high-dimensional data 
in an unsupervised way. 
In order to use non-differentiable sparsity-inducing norms as loss functions,
we plug their proximal mappings into the automatic differentiation framework.
We demonstrate unsupervised learning of U-Net for background subtraction 
using low-rank and sparse priors. 
The U-Net can learn moving objects in a training sequence without any annotation, 
and successfully detect the foreground objects in test sequences.
\end{abstract}


\input mymath.tex

\section{Introduction}

Deep learning is a powerful technique for large-scale data analytics.
There has not, however, been established a comprehensive deep-learning methodology for exploiting
mathematically modeled domain knowledge.
Even though experts in some domain have already designed a mathematical model
that can approximately represent underlying data constructions with interpretable explanatory parameters,
there is no guideline to use such domain knowledge for deep learning.
Without the domain knowledge, data scientists have to re-invent their own model typically as
a deep neural network, and fit it to large-scale data.
As a consequence, deep neural networks have to be given prior domain knowledge
indirectly through data sets annotated by experts, and not directly by existing mathematical models.
Similar criticisms are found in \cite{Muralidhar18,Marcus18}.

In this paper, we propose to explicitly use a sparse model as loss functions for training a deep neural network.
Sparse modeling is a successful strategy for concisely explain the ingredients of high-dimensional data.
The low-rank and sparse (L+S) model is suitable for analyzing a sequence or collection of data
composed of linearly dependent and sparse features~\cite{Bouwmans14}.
In deep learning methods,
the nuclear norm (a.k.a., trace norm) and $\ell_1$ norm have been respectively used for regularizing network weights
to be linearly denpendent and sparse~\cite{Yang17,Ishikawa96}.
To the best of our knowledge, our work is the first to apply these norms
to the outputs of neural network in order to achieve unsupervised learning.
In both the uses of L+S-inducing priors, there remains a potential issue in optimzation.
Since the nuclear and $\ell_1$ norms are non-smooth functions,
the loss function with these norms cannot be well minimized by the automatic differntiation frameworks for deep learning.
We improve this by plugging proximal mappings~\cite{Moreau65prox,Donoho95,Daubechies04,Cai10,Ma11}
into the gradient-based minimization.

\begin{figure}
\setlength{\unitlength}{1cm}
\hfil $\displaystyle
  \Min_{\theta} 
\lambda_*\left\|\mtr D-\mtr S_\theta(\mtr D)\right\|_*+\lambda_1\left\|\mtr S_\theta(\mtr D)\right\|_1
$\hfil
\begin{center}
\begin{picture}(8,5)
\put(0,0){\includegraphics[width=8cm]{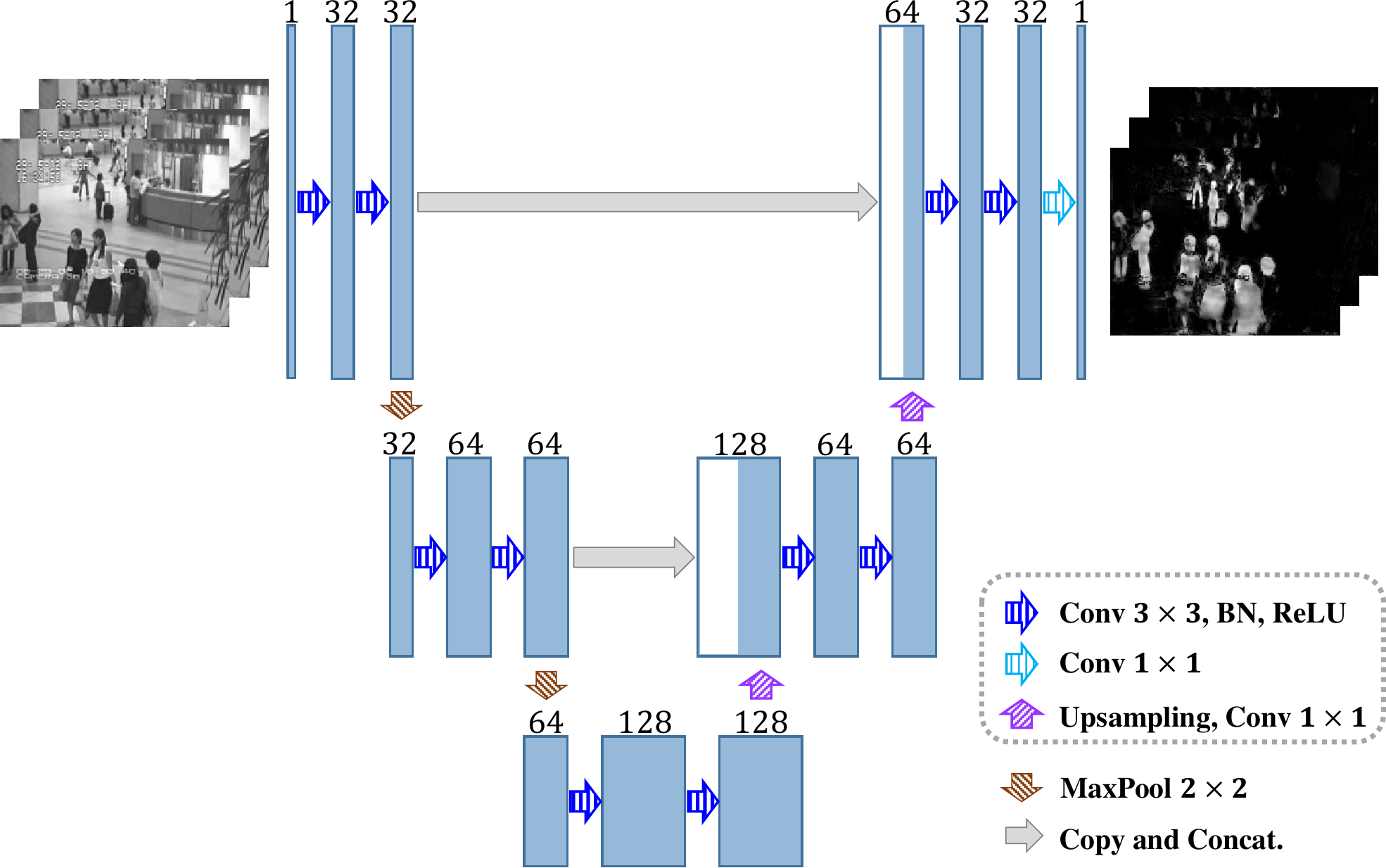}}
\put(0.7,2.8){$\mtr D$}
\put(6.7,2.8){$\mtr S_\theta(\mtr D)$}
\put(5.9,1.75){$\theta$}
\end{picture}
\end{center}
\caption{U-Net architecture (example for background subtraction).
The trainable parameters $\theta$ are optimized on training images, $\mtr D$,
so that the U-Net extracts sparse foreground objects as the output $\mtr S$
against low-rank background structure.}
\label{fig:Unet architecture}
\end{figure}
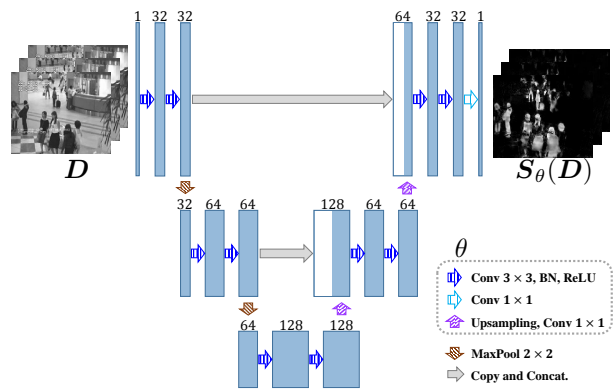

This paper begins with
an overview of deep-learning schemes to position our work.
As a typical application of unsupervised deep learning with sparsity-inducing priors,
we introduce the nuclear and $\ell_1$ norms into the loss function
for training a U-Net~\cite{Ronneberger15} as shown in Fig.~\ref{fig:Unet architecture}.
It is experimentally shown that, in a background subtraction task,
the U-Net can learn to detect foreground objects in an image sequence without any annotation.



\begin{table*}[t]
\begin{center}
\caption{Categories of network archtecture and training criteria.}
\label{tab:types}
\begin{tabular}{l|l|l}
\hline
\backslashbox{\textbf{Training}}{\textbf{Architecture}}
 & \textbf{Neural-model-based} & \textbf{Mathematical-model-based}\\
\hline\textbf{Without sparsity-inducing prior} & 
\begin{tabular}{l}
Convolutional/recurrent neural networks
\end{tabular} &
\begin{tabular}{l}
Unrolled sparse solvers,\\
e.g., LISTA\cite{Gregor10}, CORONA\cite{Solomon19}
\end{tabular}\\
\hline
\begin{tabular}{l}
\textbf{With sparsity-inducing prior}\\ \textbf{(our work)}
\end{tabular} & 
\begin{tabular}{l}
Sparse feature learning~\cite{Ng11,Jiang15,Scardapane17,Zhang18,Abavisani19,Amini19},\\
L+S regularization of network weights~\cite{Guo19}\\
\textbf{(U-Net trained with L+S output loss)}
\end{tabular} &
\begin{tabular}{l}
None (but crucial to prevent catastrophic forgetting)
\end{tabular}\\
\hline
\end{tabular}
\end{center}
\end{table*}

\section{Sparsity-aware deep-learning schemes}

As summarized in Table~\ref{tab:types},
we categorize deep-learning schemes by 
network architecture and training criteria
from the viewpoint of whether 
mathematically-modeled domain knowledge is taken into account.
There are two main approaches to designing the network architecture.
One is the successful analogy of biological neural processing (``Neural-model-based'' in Table~\ref{tab:types}):
sequential and/or recurrent connections of artificial neural network modules such as
convolutional layers and dense layers, each often followed by nonlinear activation.
Another modeling approach is the so-called unfolding or loop unrolling (``Mathematical-model-based''):
representation of a truncated iterative algorithm with a finite number of steps as
a trainable feed-forward networks.
Although there are few case studies of unfolding, 
this approach has been shown to be very beneficial when applied to sparse coding.
Gregor and LeCun~\cite{Gregor10} showed that early truncation of sparse encoders
such as the iterative soft-thresholding algorithm (ISTA)~\cite{Daubechies04}
can be trained so as to produce approximate solutions,
which impressively reduces computation time of sparse coding.
The unfolding technique was also applied to ISTA for robust PCA~\cite{Solomon19}.

There is plenty of room to incorporate sparsity-inducing priors into training as knowledge injection.
They have explicitly been utilized for sparsification of hidden unit outputs
in autoencoder-based sparse feature learning~\cite{Ng11,Jiang15,Scardapane17,Zhang18,Abavisani19,Amini19}
as well as for low-rank and/or sparse regularization of network weights~\cite{Ishikawa96,Guo19}.
The nuclear norm has not been used as a loss function yet, even though
it is {\it backpropable} 
via automatic differentiation of the singular value decomposition~\cite{Walter10}.
The $\ell_1$ norm and nuclear norm are
the convex relaxation of the $\ell_0$ norm and matrix rank, respectively.
Because of the non-smoothness of these norms,
most of the prior work mentioned above compromise some suboptimal training results
by gradient-based methods with or without smoothing the norms.
Proximal mapping as proposed in~\cite{Amini19}
is essential for the optimization of a loss function with sparsity-inducing norms,
since the optimal solution locates at a non-differentiable point in the solution space.
On fine tuning of a sparse-model-based network,
using the sparsity-inducing norms as loss functions would be crucial
to prevent catastrophic forgetting.

\section{Training U-Net for separating sparse from low-rank components}


Common occurrence of some features in a sequence or collection of data
are likely derived from normal events or spatio-temporal structure.
Sparse outlying features indicate unusual/abnormal events.
A surveilance video consists of, for example, images of stationary background with temporary foreground objects superimposed.
The background images with monotonous variation can be assumed to be linearly dependent.
The foreground objects can be sparse in time and space.

Let $\{\vec d^{(j)}\}$ ($j=1,\dots,n$) be a sequence of $m$-dimensional data or feature vectors.
We assume that $\{\vec d^{(j)}\}$ is a mixture of sequences $\{\vec l^{(j)}\}$ and $\{\vec s^{(j)}\}$,
derived respectively from usual and unusual events (e.g., sequences of background and foreground images).
The linear dependence of $\{\vec l^{(j)}\}$ can be quantified as the low-rankness of
a matrix $\mtr L=[\vec l^{(1)},\dots,\vec l^{(n)}]\in\mathbb{R}^{m\times n}$.
The sparsity of $\{\vec s^{(j)}\}$ can be evaluated as the number of nonzero entries of
$\mtr S=[\vec s^{(1)},\dots,\vec s^{(n)}]\in\mathbb{R}^{m\times n}$.

For a given matrix $\mtr D=[\vec d^{(1)},\dots,\vec d^{(n)}]\in\mathbb{R}^{m\times n}$,
its representation as a sum of the low-rank matrix $\mtr L$ and the sparse matrix $\mtr S$
is called the low-rank and sparse (L+S) model.
Fitting the L+S model to $\mtr D$ can be posed as the following minimization problem.
\begin{equation}
 \Min_{(\mtr L,\mtr S)} \frac{1}{2}\|\mtr D-(\mtr L+\mtr S)\|_F^2
+\lambda_*\|\mtr L\|_*+\lambda_1\|\mtr S\|_1
\label{eq:L+S fitting}
\end{equation}
Here, $\|\cdot\|_F$ indicates the Frobenius norm.
Minimizing the nuclear norm $\|\mtr L\|_*$ and $\ell_1$ norm $\|\mtr S\|_1$
promotes the low-rankness and sparseness of the matrices $\mtr L$ and $\mtr S$, respectively.
Their balance is controled by the positive constants $\lambda_*$ and $\lambda_1$.
The minimization problem in Eq.~(\ref{eq:L+S fitting}) is rewritten as
\begin{equation}
 \Min_{\mtr X} \frac{1}{2}\|\mtr D-\mtr A\mtr X\|_F^2+g(\mtr X)
\label{eq:two convex functions}
\end{equation}
where $\mtr X=[\mtr L^\top,\mtr S^\top]^\top\in\mathbb{R}^{2m\times n}$ and
$\mtr A=[\mtr I,\mtr I]\in\mathbb{R}^{m\times 2m}$ with the $m\times m$ identity matrix $\mtr I$.
The scalar function
\[
 g:\bmatrix{c}{\mtr L\\ \mtr S}\mapsto
\lambda_*\|\mtr L\|_*+\lambda_1\|\mtr S\|_1
\]
is a non-smooth convex function.
Proximal forward-backward splitting algorithm~\cite{Combettes05} is a reliable approach to
the solution of the problem in Eq.~(\ref{eq:two convex functions}).
Until convergence, the algorithm takes the iterative steps described as
\begin{equation}
 \mtr X^{(k+1)}=\prox_{\alpha_k g}(\mtr X^{(k)}+\alpha_k\mtr A^\top(\mtr D-\mtr A\mtr X^{(k)})).
\label{eq:proximal forward-backward}
\end{equation}
The superscript $\top$ denotes the matrix transposition.
$\alpha_k$ must be positive and not greater than $1/\kappa_{\max}^2=1/2$
where $\kappa_{\max}$ is the largest singular value of $\mtr A$.
The proximal mapping~\cite{Moreau65prox} of a matrix $\mtr Q$ with respect to a convex function $f$ is defined as
\[
 \prox_f(\mtr Q)=\argmin_{\mtr X}f(\mtr X)+\frac{1}{2}\|\mtr X-\mtr Q\|_F^2.
\]
The proximal forward-backward steps in Eq.~(\ref{eq:proximal forward-backward})
are explicitly written in the following forms.
\begin{eqnarray}
\mtr L^{(k+1)}&=&\textrm{svt}(\mtr L^{(k)}+\alpha(\mtr D-\mtr L^{(k)}-\mtr S^{(k)}),\alpha\lambda_*)\label{eq:update L}\\
\mtr S^{(k+1)}&=&\soft(\mtr S^{(k)}+\alpha(\mtr D-\mtr L^{(k)}-\mtr S^{(k)}),\alpha\lambda_1)\label{eq:update S}
\end{eqnarray}
The singular value thresnolding operation~\cite{Cai10,Ma11}, denoted as svt, of a matrix $\mtr Q$ is defined as
\[
 \textrm{svt}(\mtr Q,\tau)=\mtr U^\top\soft(\mtr K,\tau)\mtr V^\top.
\]
Here, the matrices $\mtr U$, $\mtr K$ and $\mtr V$ are the singular value decomposition (SVD) of $\mtr Q$.
The soft thresholding operation~\cite{Donoho95,Daubechies04}, denoted as soft, is defined as
\[
 \soft(q,\tau)=\sign(q)\max(|q|-\tau,0),
\]
and it works element-wise on matrices.

If $\{\vec d^{(j)}\}$ is a collection of images with (L+S) structure,
hourglass network architecture with convolutional layers,
such as the U-Net~\cite{Ronneberger15}, is a best practice
for esitmating either or both of the low-rank and sparse features.
In order to roughly train a U-Net model that
outputs the sparse component $\vec s^{(j)}$ for the input $\vec d^{(j)}$,
one can approximately minimize
\begin{equation}
\textrm{loss}(\theta)
=\lambda_*\|\mtr D-\mtr S_\theta(\mtr D)\|_*
+\lambda_1\|\mtr S_\theta(\mtr D)\|_1
\label{eq:U-Net loss for adam}
\end{equation}
with respect to the set of network weights, $\theta$, using
an automatic differentiation framework for deep learning, e.g., PyTorch~\cite{PyTorch19etal}.
Some of such frameworks provide a {\it backpropable} SVD function, and enable us
to implement the nuclear norm in Eq.~(\ref{eq:U-Net loss for adam}).

We can further polish the U-Net model.
Analogous to Eqs.~(\ref{eq:update L}) and (\ref{eq:update S}),
we repeat the following proximal mapping procedure.
\begin{itemize}
\item  Take the copies of the current estimates:
\begin{eqnarray*}
\mtr L &\gets&\mtr D-\mtr S_\theta(\mtr D)\\
\mtr S &\gets&\mtr S_\theta(\mtr D).
\end{eqnarray*}
\item Update $\theta$: take a single step of gradient descent by backpropagation from
\begin{eqnarray}
&&\lambda_*\|\mtr L+\alpha(\mtr S-\mtr S_\theta(\mtr D)\|_*\nonumber\\
&&
\quad{}+\lambda_1\|\mtr S_\theta(\mtr D)+\alpha(\mtr S-\mtr S_\theta(\mtr D)\|_1
\label{eq:U-Net loss for GD}
\end{eqnarray}
using $\mtr Q-\textrm{svt}(\mtr Q,\tau)$ and $\mtr Q-\soft(\mtr Q,\tau)$
as the gradients of the nuclear and $\ell_1$ norms at $\mtr Q$ on the backward pass, respectively.
\end{itemize}

%

\section{Experiment}

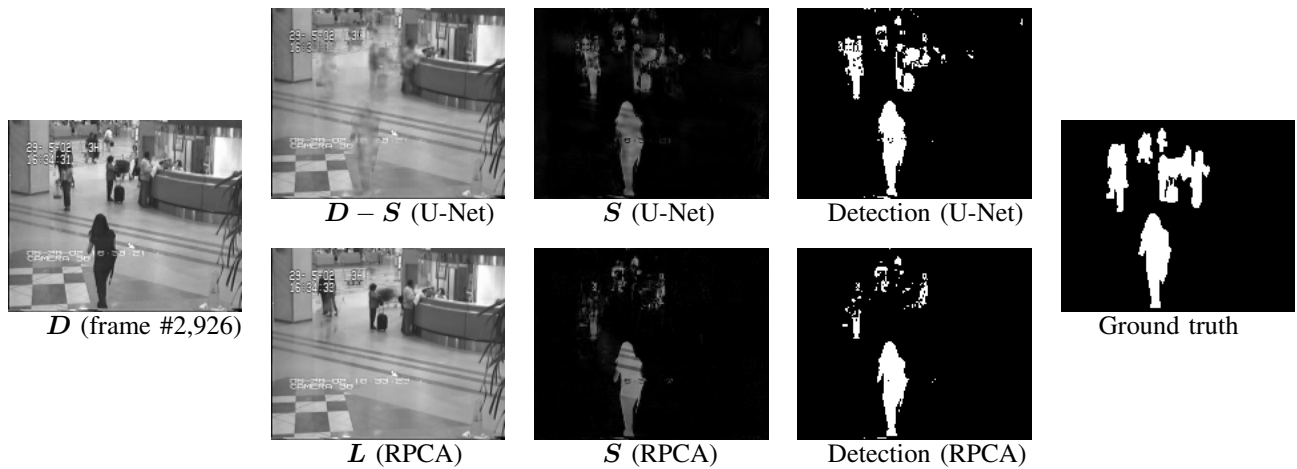
\begin{figure*}[t]
\setlength{\unitlength}{1cm}
\begin{center}
\begin{picture}(17.2,6)
\put(0,2){\includegraphics[scale=0.5]{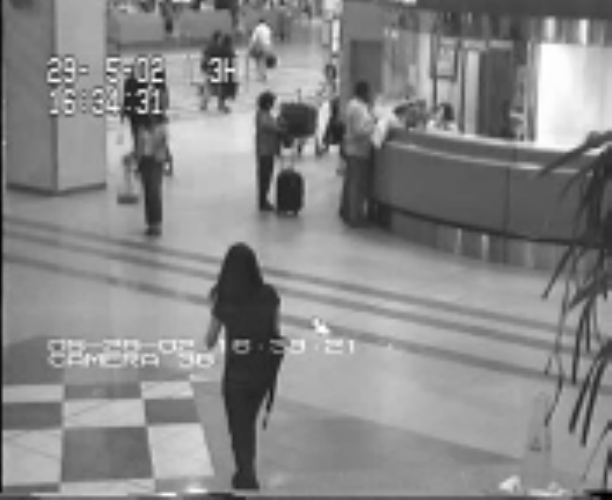}}
\put(3.5,3.5){\includegraphics[scale=0.5]{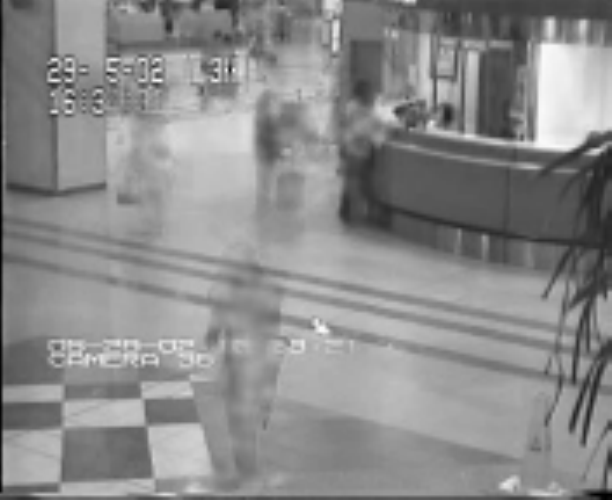}}
\put(7,3.5){\includegraphics[scale=0.5]{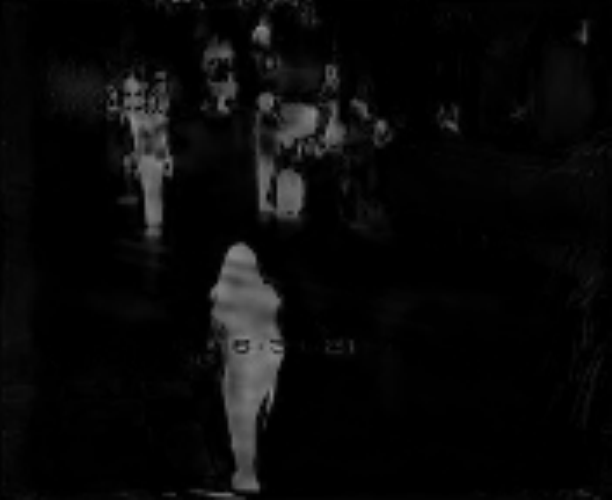}}
\put(10.5,3.5){\includegraphics[scale=0.5]{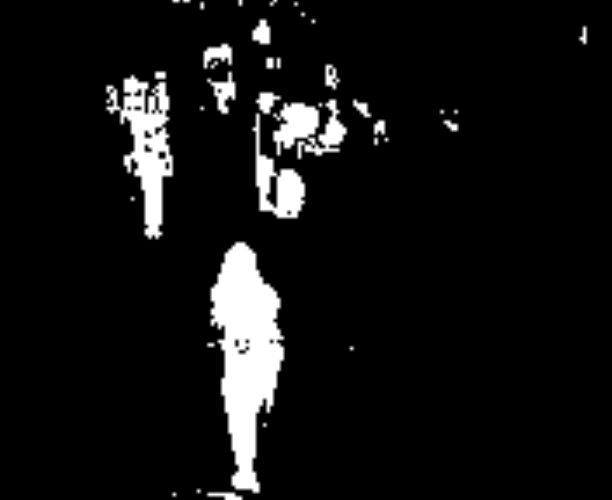}}
\put(3.5,0.3){\includegraphics[scale=0.5]{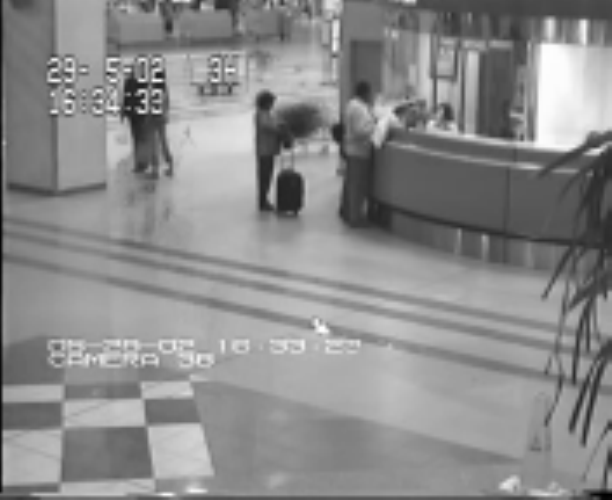}}
\put(7,0.3){\includegraphics[scale=0.5]{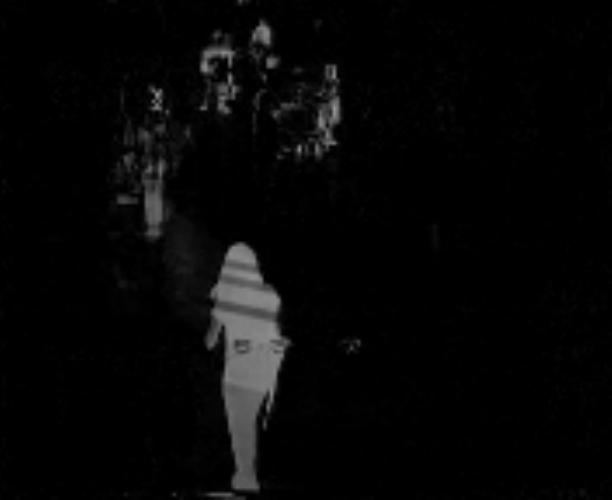}}
\put(10.5,0.3){\includegraphics[scale=0.5]{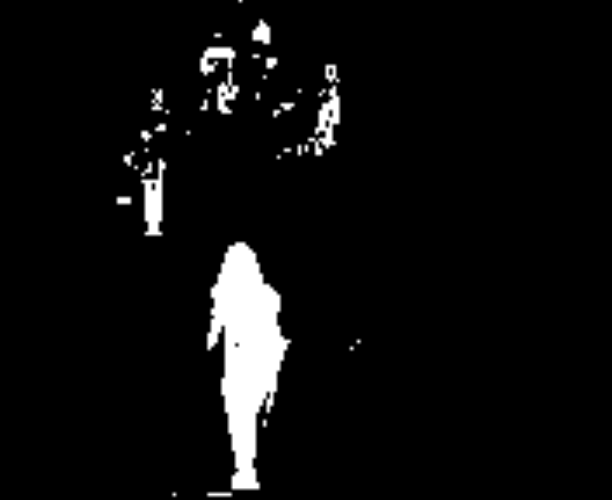}}
\put(14,2){\includegraphics[scale=0.5]{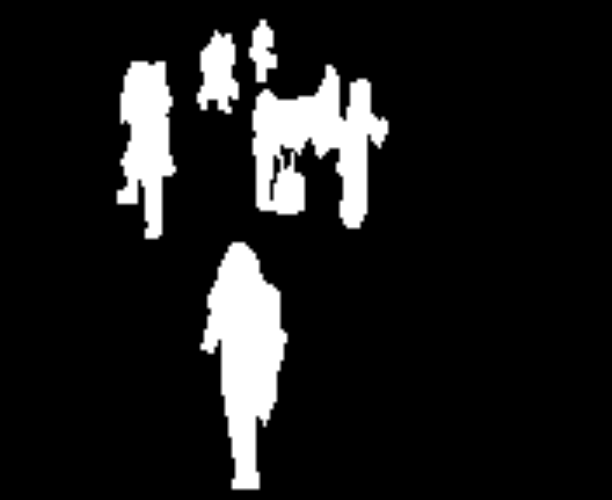}}
\put(0.5,1.7){$\mtr D$ (frame \#2{,}926)}
\put(14.5,1.7){Ground truth}
\put(4.2,3.2){$\mtr D-\mtr S$ (U-Net)}
\put(7.9,3.2){$\mtr S$ (U-Net)}
\put(10.9,3.2){Detection (U-Net)}
\put(4.5,0){$\mtr L$ (RPCA)}
\put(7.9,0){$\mtr S$ (RPCA)}
\put(10.9,0){Detection (RPCA)}
\end{picture}
\end{center}
\caption{Background subraction via U-Net (first row) and robust PCA (second row).
U-Net is trained with 50 images in this scene.
This video frame \#2{,}926 is not included in training.
The result by robust PCA is obtained from 50 images around this frame.}
\label{fig:airport 2926}
\end{figure*}

We demonstrate the unsupervised learning of U-Net via its application to background subtraction
on an image sequence ``airport''~\cite{Li04}.
The sequence consists of 3{,}584 color images (\#1{,}000 to \#4{,}583) of size 144$\times$176.
We convert the colors to grayscale values ranging from 0 to 1, and use randomly chosen 50 frames for training.
U-Net treats the training images as a tensor of size 50$\times$1$\times$144$\times$176.
When evaluating the losses in Eqs.~(\ref{eq:U-Net loss for adam}) and (\ref{eq:U-Net loss for GD}),
the input and output tensors of U-Net are arranged into the matrices $\mtr D$ and $\mtr S_\theta(\mtr D)$
with $m=$144$\times$176$=$25{,}344 dimensional $n=$50 column vectors of pixel values of corresponding images.

The U-Net architecture is shown in Fig.~\ref{fig:Unet architecture}.
Each blue box corresponds to a multi-channel feature map.
The number of channels is denoted on top of the box.
White boxes represent copied feature maps.
Each convolutional layer with 3$\times$3 filters is followed by
a batch normalization (BN) layer~\cite{Ioffe15} and a rectified linear unit (ReLU).
The number of trainable params are 518{,}433, which roughly amounts to 2GB,
and 8GB GPU memory are enough for forward and backward pass.
We trained the U-Net as described in the previous section.
We set $\lambda_*=1$, $\lambda_1=5\times 10^{-3}$, $\alpha=0.5$.
We ran the Adam optimizer~\cite{Diederik15} with a learning rate of $3\times 10^{-4}$ for 2{,}000 epochs
to decrease the loss in Eq.~(\ref{eq:U-Net loss for adam}).
Then, we repeated the proximal mapping precedure 3{,}000 times
to decrease the loss in Eq.~(\ref{eq:U-Net loss for GD}) with a learning rate of $3\times 10^{-8}$.

Figure~\ref{fig:airport 2926} 
shows an example of the background subtraction result
for a test image in the sequence.
For comparison, the low-rank and sparse components of 50 images around the test image frame
are shown in Fig.~\ref{fig:airport 2926} as the result of robust principal component analysis (RPCA).
The RPCA result can be obtained by repeating the proximal forward-backward steps as shown
in Eqs.~(\ref{eq:update L}) and (\ref{eq:update S}) until convergence.
One can see that RPCA misses foreground objects staying 
within this time interval, and represents them in the low-rank component (see persons in $\mtr L$).
Including such objects, the trained U-Net detects the foregrounds better than RPCA,
even when a pedestrian is partially occluded by the superimposed timestamp on the upper left side in the frame.
This comparison implies that the U-Net learns apparent shapes of foreground objects
withtout any annotation.
We identified the foreground pixels by the Otsu binarization of the absolute values of $\mtr S$.
The precision and recall of this U-Net output are
95.8\% and 63.2\% while those by RPCA are 96.5\% and 49.6\%, respectively.
As another advantage, we would like to note that
the forward computation of trained U-Net is much faster than applying RPCA
to any batch of sequential images.

\section{Concluding remarks}

Sparsity-inducing priors, the low-rank and sparse priors in particular,
are very worth incorporating in deep learning.
We illustrated, in a background subtraction task,
how to inject sparse prior information into a deep network.
Although the U-Net in the experiment may overfit the scene,
we expect that it will generalize well
if it experiences similar foreground objects in various training sequences.
Our unsupervised deep learning would be advantageous when applied to
medical images with L+S structure, e.g.,
angiography under free-breathing condition~\cite{Kawabe19tsakai}.

\bibliographystyle{IEEEtran}
\bibliography{mybib_dl,mybib_sparse,mybib_pca,mybib_misc,mybib_mi}

\end{document}

%% file: mymath.tex
%
%

\def\refeq#1{(\ref{eq:#1})}
\def\tten#1{{\!\times\!10^{#1}}}
\def\mit#1{{\mathit #1}}
\def\confrac#1#2{%
 \frac{\displaystyle{%
  \strut\hfill{#1}\hfill\;\vrule}}%
   {\displaystyle{%
    \strut\vrule\;\hfill{#2}\hfill}}}

\def\sfrac#1#2{{\begingroup#1\endgroup\over#2}}

\def\sub#1{${}_{\mathrm{#1}}$}
\def\sup#1{${}^{\mathrm{#1}}$}
\def\msub#1{_{\mbox{\scriptsize #1}}}
\def\msup#1{^{\mbox{\scriptsize #1}}}

\def\rnum#1{\expandafter{\romannumeral #1}}
\def\Rnum#1{\uppercase\expandafer{\romannumeral #1}}

\def\bmatrix#1#2{\left[\begin{array}{#1}#2\end{array}\right]}
\def\smatrix#1#2#3{\mbox{#1$\bmatrix{#2}{#3}$}}

\def\vec#1{\boldsymbol{#1}}

\def\est#1{\hat{#1}}
\def\tru#1{#1^\circ}
\def\amin#1{#1^\star}
\def\prx#1{\hat{#1}}
\def\stp#1{#1^\star}
\def\old#1{{#1'}}
\def\nnz{s}

\def\mtr#1{\boldsymbol{#1}}

\def\complement#1{\overline{#1}}

\def\Event#1{\mathrm{#1}}

\def\mplus{\boldsymbol{+}}

\def\defas{:=}
\def\subs{\leftarrow}
\def\lap{{\rm\Delta}}
\def\Omicron{\mathrm{O}}
\def\omicron{\mathrm{o}}

\def\Obj{F}
\def\Pbj{J}
\def\Maj{Q}

\def\iff{\quad\Leftrightarrow\quad}

\def\onl#1{\operatornamewithlimits{\mathrm{#1}}}
\def\Min{\mathop{\textrm{Minimize }}\limits}
\def\Max{\mathop{\textrm{Maximize }}\limits}
\newcommand{\argmin}{\operatornamewithlimits{\mathrm{arg\,min}}}
\newcommand{\Argmin}{\operatornamewithlimits{\mathrm{Arg\,min}}}
\newcommand{\argmax}{\operatornamewithlimits{\mathrm{arg\,max}}}
\newcommand{\Argmax}{\operatornamewithlimits{\mathrm{Arg\,max}}}

\def\sup{\mathop{\textrm{sup}}\limits}
\def\inf{\mathop{\textrm{inf}}\limits}

\def\vectorize{\mathop{\textrm{vec}}\nolimits}
\def\unvectorize{\mathop{\textrm{mat}}\nolimits}

\def\Schatten#1{\mathop{S}_#1}

\def\supp{\mathop{\textrm{supp}}\nolimits}
\def\spark{\mathop{\textrm{spark}}\nolimits}
\def\soft{\mathop{\textrm{soft}}\nolimits}
\def\hard{\mathop{\textrm{hard}}\nolimits}
\def\prox{\mathop{\textrm{prox}}\nolimits}
\def\epi{\mathop{\textrm{epi}}\nolimits}

\def\cond{\mathop{\textrm{cond}}\nolimits}
\def\rank{\mathop{\textrm{rank}}\nolimits}
\def\diag{\mathop{\textrm{diag}}\nolimits}
\def\tr{\mathop{\textrm{tr}}\nolimits}
\def\trace{\mathop{\textrm{trace}}\nolimits}
\def\Kernel{\mathop{\textrm{Kernel}}\nolimits}
\def\kernel{\mathop{\textrm{kernel}}\nolimits}
\def\nullsp{\mathop{\textrm{null}}\nolimits}
\def\im{\mathop{\textrm{im}}\nolimits}
\def\Span{\mathop{\textrm{span}}\nolimits}
\def\Sparse{\mathop{\textrm{Sparse}}\nolimits}
\def\sparse{\mathop{\textrm{sparse}}\nolimits}
\def\colsp{\mathop{\textrm{colsp}}\nolimits}

\def\Sgn{\mathop{\textrm{Sgn}}\nolimits}
\def\sgn{\mathop{\textrm{sgn}}\nolimits}
\def\sign{\mathop{\textrm{sign}}\nolimits}
\def\Expect{\mathop{\rm{E}}\nolimits}
\def\Prob{\mathbb{P}}
\def\Var{\mathop{\rm{Var}}\nolimits}
\def\Comb#1#2{{}_{#1}\mbox{C}_{#2}}

\def\Fourier{\mathop{\mathscr{F}}\nolimits}

\def\st{\mathop{\quad\textrm{subject to}\quad}\nolimits}

\newfont{\bg}{cmr10 scaled\magstep4}
\def\bigzerou{\smash{\hbox{\bg 0}}}
\def\bigzerol{\smash{\lower1.7ex\hbox{\bg 0}}}
\def\Re{\mathop{\mathcal R\!e}\nolimits}
\def\Im{\mathop{\mathcal I\!m}\nolimits}
\def\D{{\mit\Delta}}
\def\d{\mbox{d}}
\def\norm#1{\left\|#1\right\|}

\def\dg{^{\circ}}
\def\deg{^{\circ}}

\def\shrtstckl#1{\shortstack[l]{#1\\{}}}